\documentclass[11pt,a4paper]{article}
\usepackage{authblk}
\PassOptionsToPackage{hyphens}{url}\usepackage[hyperref]{emnlp2018}
\usepackage{times}
\usepackage{siunitx}
\usepackage{latexsym}
\usepackage{microtype}
\usepackage{enumitem}

\aclfinalcopy 

\title{Marrying Universal Dependencies and Universal Morphology}

\author[1]{Arya D. McCarthy}
\author[2]{Miikka Silfverberg}
\author[1]{Ryan Cotterell}
\author[2]{\authorcr Mans Hulden}
\author[1]{David Yarowsky}
\affil[1]{Johns Hopkins University}
\affil[2]{University of Colorado Boulder}
\affil[ ]{\{\texttt{arya,rcotter2,yarowsky}\}\texttt{@jhu.edu}}
\affil[ ]{\{\texttt{miikka.silfverberg,mans.hulden}\}\texttt{@colorado.edu}}

\date{}

\usepackage{microtype}  
\usepackage{booktabs}  
\usepackage{todonotes}
\usepackage{dirtytalk}  
\usepackage{graphicx}  

\def\Snospace~{\S{}}

\newcommand{\tag}[1]{\textsc{#1}}
\newcommand{\term}[1]{\textbf{#1}}
\newcommand{\lemma}[1]{\texttt{#1}}
\newcommand{\form}[1]{\say{#1}}

\begin{document}
\maketitle
\begin{abstract}
The Universal Dependencies (UD) and Universal Morphology (UniMorph) projects each present schemata for annotating the morphosyntactic details of language.
Each project also provides corpora of annotated text in many languages---UD at the token level and UniMorph at the type level.
As each corpus is built by different annotators, language-specific decisions hinder the goal of universal schemata.
With compatibility of tags, each project's annotations could be used to validate the other's.
Additionally, the availability of both type- and token-level resources would be a boon to tasks such as parsing and homograph disambiguation.
To ease this interoperability, we present a deterministic mapping from Universal Dependencies~v2 features into the UniMorph schema.
We validate our approach by lookup in the UniMorph corpora and find a macro-average of \(64.13\%\) recall.
We also note incompatibilities due to paucity of data on either side. Finally, we present a critical evaluation of the foundations, strengths, and weaknesses of the two annotation projects. 
\end{abstract}

\section{Introduction}

\begin{figure}
\includegraphics[width=\linewidth]{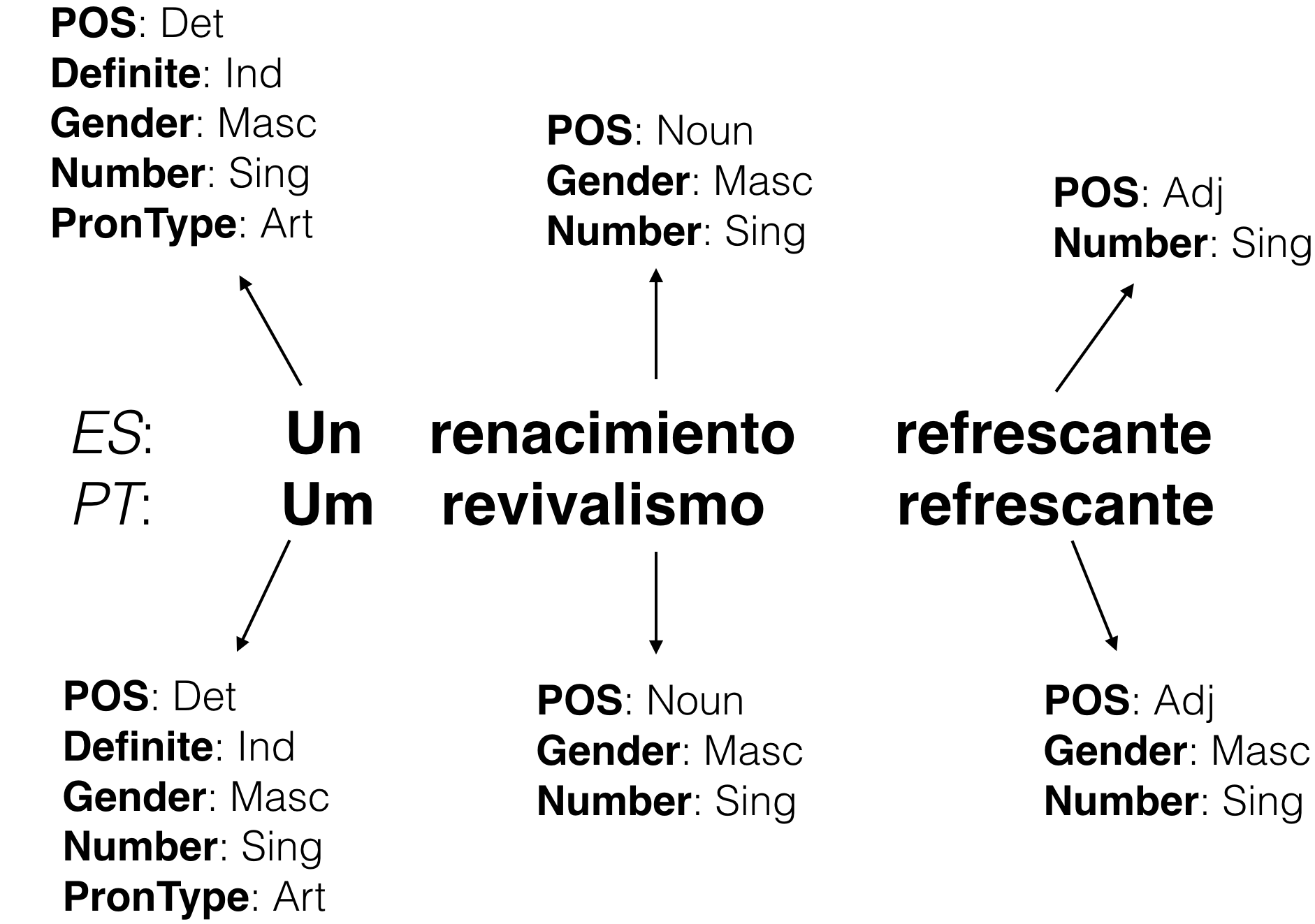}
\caption{Example of annotation disagreement in UD between two languages on translations of one phrase, reproduced from \citet{malaviya2018neural}. The final word in each, \form{\emph{refrescante}}, is not inflected for gender: It has the same surface form whether masculine or feminine. Only in Portuguese, it is annotated as masculine to reflect grammatical concord with the noun it modifies.}
\label{fig:disagreement}
\end{figure}

The two largest standardized, cross-lingual datasets for morphological annotation are provided by the Universal Dependencies \citep[UD;][]{nivre2017universal} and Universal Morphology \cite[UniMorph;][]{sylakglassman2015,kirov2018unimorph} projects.
Each project's data are annotated according to its own cross-lingual schema, prescribing how features like gender or case should be marked.
The schemata capture largely similar information, so one may want to leverage both UD's token-level treebanks and UniMorph's type-level lookup tables and unify the two resources.
This would permit a leveraging of both the token-level UD treebanks and the type-level UniMorph tables of paradigms.
Unfortunately, neither resource perfectly realizes its schema.
On a dataset-by-dataset basis, they incorporate annotator errors, omissions, and human decisions when the schemata are underspecified; one such example is in \autoref{fig:disagreement}.

A dataset-by-dataset problem demands a dataset-by-dataset solution; our task is not to translate a \emph{schema}, but to translate a \emph{resource}.
Starting from the idealized schema, we create a rule-based tool for converting UD-schema annotations to UniMorph annotations, incorporating language-specific post-edits that both correct infelicities and also increase harmony between the datasets themselves (rather than the schemata).
We apply this conversion to the 31 languages with both UD and UniMorph data, and we report our method's recall, showing an improvement over the strategy which just maps corresponding schematic features to each other.
Further, we show similar downstream performance for each annotation scheme in the task of morphological tagging.

This tool enables a synergistic use of UniMorph and Universal Dependencies, as well as teasing out the annotation discrepancies within and across projects. When one dataset disobeys its schema or disagrees with a related language, the flaws may not be noticed except by such a methodological dive into the resources. When the maintainers of the resources ameliorate these flaws, the resources move closer to the goal of a universal, cross-lingual inventory of features for morphological annotation.

The contributions of this work are:
\begin{itemize}
\item We detail a deterministic mapping from UD morphological annotations to UniMorph. Language-specific edits of the tags in 31 languages increase harmony between converted UD and existing UniMorph data (\autoref{sec:conversion}).
\item We provide an implementation of this mapping and post-editing, which replaces the UD features in a CoNLL-U file with UniMorph features.\footnote{Available at \url{https://www.github.com/unimorph/ud-compatibility}.}
\item We demonstrate that downstream performance tagging accuracy on UD treebanks is similar, whichever annotation schema is used~(\autoref{sec:results}).
\item We provide a partial inventory of missing attributes or annotation inconsistencies in both UD and UniMorph, a guidepost for strengthening and harmonizing each resource.
\end{itemize}

\section{Background: Morphological Inflection}

Morphological \term{inflection} is the act of altering the base form of a word (the \term{lemma}, represented in \lemma{fixed-width type}) to encode morphosyntactic features. As an example from English, \lemma{prove} takes on the \term{form} \form{proved} to indicate that the action occurred in the past. (We will represent all surface forms in quotation marks.) The process occurs in the majority of the world's widely-spoken languages, typically through meaningful affixes. The breadth of forms created by inflection creates a challenge of data sparsity for natural language processing: The likelihood of observing a particular word form diminishes.

A classic result in psycholinguistics \citep{berko1958child}
shows that inflectional morphology is a fully productive process. Indeed,
it cannot be that humans simply have the equivalent of a lookup table, where they store the inflected forms for retrieval as the syntactic context requires. Instead, there needs to be a mental process that can generate properly inflected
words on demand. \citet{berko1958child} showed this insightfully through
the \say{wug}-test, an experiment where she forced
participants to correctly inflect out-of-vocabulary lemmata,
such as the novel noun \lemma{wug}.

Certain features of a word do not vary depending on its context: In German or Spanish where nouns are gendered, the word for \lemma{onion} will always be grammatically feminine. Thus, to prepare for later discussion, we divide the morphological features of a word into two categories: the modifiable \term{inflectional features} and the fixed \term{lexical features}.

A \term{part of speech (POS)} is a coarse syntactic category (like \say{verb}) that begets a word's particular menu of lexical and inflectional features. In English, verbs express no gender, and adjectives do not reflect person or number. The part of speech dictates a set of inflectional \term{slots} to be filled by the surface forms. Completing these slots for a given lemma and part of speech gives a \term{paradigm}: a mapping from slots to surface forms. Regular English verbs have five slots in their paradigm \citep{long1957paradigms}, which we illustrate for the verb \lemma{prove}, using simple labels for the forms in \autoref{tab:ptb}.

\begin{table}
    \centering
    \begin{tabular}{l l l}
        \toprule
        Simple label & Form & PTB tag \\
        \midrule
        Present, 3rd singular & \form{proves} & VBZ \\
        Present, other & \form{prove} & VBP \\
        Past & \form{proved} & VBD \\
        Past participle & \form{proven} & VBN \\
        Present participle & \form{proving} & VBG \\
        \bottomrule
    \end{tabular}
    \caption{Inflected forms of the English verb \lemma{prove}, along with their Penn Treebank tags}
    \label{tab:ptb}
\end{table}

A morphosyntactic \term{schema} prescribes how language can be annotated---giving stricter categories than our simple labels for \lemma{prove}---and can vary in the level of detail provided. Part of speech tags are an example of a very coarse schema, ignoring details of person, gender, and number. A slightly finer-grained schema for English is the Penn Treebank tagset \cite{marcus1993building}, which includes signals for English morphology. For instance, its \tag{VBZ} tag pertains to the specially inflected 3rd-person singular, present-tense verb form (e.g.\ \form{proves} in \autoref{tab:ptb}).

If the tag in a schema is detailed enough that it exactly specifies a slot in a paradigm, it is called a \term{morphosyntactic description (MSD)}.\footnote{Other sources will call this a morphological tag or bundle. We avoid the former because of the analogy to POS tagging; a morphological tag is not atomic.} These descriptions require varying amounts of detail: While the English verbal paradigm is small enough to fit on a page, the verbal paradigm of the Northeast Caucasian language Archi can have over~\SI{1500000} slots~\citep{kibrik1998archi}.%

\section{Two Schemata, Two Philosophies}

Unlike the Penn Treebank tags, the UD and UniMorph schemata are cross-lingual and include a fuller lexicon of attribute-value pairs, such as \tag{\textbf{Person}:~1}. Each was built according to a different set of principles. UD's schema is constructed bottom-up, adapting to include new features when they're identified in languages. UniMorph, conversely, is top-down: A cross-lingual survey of the literature of morphological phenomena guided its design. UniMorph aims to be linguistically complete, containing all known morphosyntactic attributes. Both schemata share one long-term goal: a total inventory for annotating the possible morphosyntactic features of a word.

\subsection{Universal Dependencies}

\begin{table*}[t]
    \centering
    \begin{tabular}{l l}
        \toprule
        Schema & Annotation \\
        \midrule
        UD & \tag{VERB\qquad{}Mood=Ind\textbar{}Number=Sing\textbar{}Person=3\textbar{}Tense=Imp\textbar{}VerbForm=Fin} \\
        UniMorph & \tag{V;IND;PST;1;SG;IPFV} \\
                 & \tag{V;IND;PST;3;SG;IPFV} \\
        \bottomrule
    \end{tabular}
    \caption{Attested annotations for the Spanish verb form \say{\emph{mandaba}} \say{I/he/she/it commanded}. Note that UD separates the part of speech from the remainder of the morphosyntactic description. In each schema, order of the values is irrelevant.}
    \label{tab:annotations}
\end{table*}

The Universal Dependencies morphological schema comprises part of speech and 23 additional attributes (also called features in UD) annotating meaning or syntax, as well as language-specific attributes. 
In order to ensure consistent annotation, attributes are included into the general UD schema if they occur in several corpora. Language-specific attributes are used when only one corpus annotates for a specific feature.

The UD schema seeks to balance language-specific and cross-lingual concerns. It annotates for both inflectional features such as case and lexical features such as gender. Additionally, the UD schema annotates for features which can be interpreted as derivational in some languages. For example, the Czech UD guidance uses a \tag{Coll} value for the \tag{\textbf{Number}} feature to denote mass nouns (for example, "{\it lidstvo}" "humankind" from the root "{\it lid}" "people").\footnote{Note that \tag{\textbf{Number}: Coll} does not actually figure in the Czech corpus.}

UD represents a confederation of datasets \citep[see, e.g.,][]{dirix2017universal} annotated with dependency relationships (which are not the focus of this work) and morphosyntactic descriptions. Each dataset is an annotated treebank, making it a resource of \term{token-level} annotations. The schema is guided by these treebanks, with feature names chosen for relevance to native speakers. (In \autoref{sec:unimorph}, we will contrast this with UniMorph's treatment of morphosyntactic categories.) The UD datasets have been used in the CoNLL shared tasks \citep[2018 to appear]{zeman2017conll}.

\subsection{UniMorph} \label{sec:unimorph}

In the Universal Morphological Feature Schema \citep[UniMorph schema,][]{sylak2016composition}, there are at least 212 values, spread across 23 attributes. It identifies some attributes that UD excludes like information structure and deixis, as well as providing more values for certain attributes, like 23 different noun classes endemic to Bantu languages. As it is a schema for marking morphology, its part of speech attribute does not have POS values for punctuation, symbols, or miscellany (\tag{Punct}, \tag{Sym}, and~\tag{X} in Universal Dependencies).

Like the UD schema, the decomposition of a word into its lemma and MSD is directly comparable across languages. Its features are informed by a distinction between \term{universal categories}, which are widespread and psychologically \say{real} to speakers; and \term{comparative concepts}, only used by linguistic typologists to compare languages \citep{haspelmath2010comparative}. Additionally, it strives for identity of meaning across languages, not simply similarity of terminology. As a prime example, it does not regularly label a dative case for nouns, for reasons explained in depth by \citet{haspelmath2010comparative}.\footnote{\say{The Russian Dative, the Korean Dative, and the Turkish Dative are similar enough to be called by the same name, but there are numerous differences between them and they cannot be simply equated with each other. Clearly, their nature is not captured satisfactorily by saying that they are instantiations of a crosslinguistic category \say{dative}.} \citep{haspelmath2010comparative}}

The UniMorph resources for a language contain complete paradigms extracted from Wiktionary \citep{kirov2016very, kirov2018unimorph}. Word \term{types} are annotated to form a database, mapping a lemma--tag pair to a surface form. The schema is explained in detail in \citet{sylak2016composition}. It has been used in the SIGMORPHON shared task \citep{cotterell2016sigmorphon} and the CoNLL--SIGMORPHON shared tasks \citep{cotterell2017conll, cotterell-EtAl:2018}. Several components of the UniMorph schema have been adopted by UD.%
\footnote{\url{http://universaldependencies.org/v2/features.html\#comparison-with-unimorph}}

\subsection{Similarities in the annotation}

While the two schemata annotate different features, their annotations often look largely similar. Consider the attested annotation of the Spanish word \form{\emph{mandaba}} \say{(I/he/she/it) commanded}. \autoref{tab:annotations} shows that these annotations share many attributes.

Some conversions are straightforward: \tag{VERB} to \tag{V}, \tag{Mood=Ind} to \tag{IND}, \tag{Number=Sing} to \tag{SG}, and \tag{Person=3} to \tag{3}.%
\footnote{The curious reader may wonder why there are two rows of UniMorph annotation for \say{\emph{mandaba}}, each with a different recorded person. The word displays \textbf{syncretism}, meaning that a single form realizes multiple MSDs. UniMorph chooses to mark these separately for the sake of its decomposable representation. As this ambiguity is systematic and pervasive in the language, one can imagine a unified paradigm slot \tag{V;IND;PST;\{1/3\};SG;IPFV} \citep{baerman2005syntax}.}
One might also suggest mapping \tag{Tense=Imp} to \tag{IPFV}, though this crosses semantic categories: \tag{IPFV} represents the imperfective \emph{aspect}, whereas \tag{Tense=Imp} comes from \term{imperfect}, the English name often given to Spanish's \emph{pasado continuo} form. The imperfect is a verb form which combines both past tense and imperfective aspect. UniMorph chooses to split this into the atoms \tag{PST} and \tag{IPFV}, while UD unifies them according to the familiar name of the tense.

\section{UD treebanks and UniMorph tables} \label{sec:resources}

Prima facie, the alignment task may seem trivial. But we've yet to explore the humans in the loop. This conversion is a hard problem because we're operating on idealized schemata. We're actually annotating human decisions---and human mistakes. If both schemata were perfectly applied, their overlapping attributes could be mapped to each other simply, in a cross-lingual and totally general way. Unfortunately, the resources are imperfect realizations of their schemata. The cross-lingual, cross-resource, and within-resource problems that we'll note mean that we need a tailor-made solution for each language.

Showcasing their schemata, the Universal Dependencies and UniMorph projects each present large, annotated datasets. UD's v2.1 release \citep{nivre2017universal} has 102 treebanks in 60 languages. The large resource, constructed by independent parties, evinces problems in the goal of a universal inventory of annotations. Annotators may choose to omit certain values (like the coerced gender of \emph{refrescante} in \autoref{fig:disagreement}), and they may disagree on how a linguistic concept is encoded. (See, e.g., \citeauthor{haspelmath2010comparative}'s (\citeyear{haspelmath2010comparative}) description of the dative case.) Additionally, many of the treebanks \say{were created by fully- or semi-automatic conversion from treebanks with less comprehensive annotation schemata than UD} \citep{malaviya2018neural}. For instance, the Spanish word \say{\emph{vas}} \say{you go} is incorrectly labeled \tag{\textbf{Gender:} Fem\textbar{}\textbf{Number:} Pl} because it ends in a character sequence which is common among feminine plural nouns. (Nevertheless, the part of speech field for \say{\emph{vas}} is correct.)

\begin{figure*}
\centering
\begin{tabular}{l l l l l l l}
\toprule
tegarg & \textbf{latme}-ye & bad-i & be & ba:q-e & man & \textbf{zad}. \\
Hail & damage-\tag{EZ} & bad-\tag{INDEF PAR} & to & garden-\tag{EZ} & \tag{1.s} & beat-\tag{PST}. \\
\midrule
\multicolumn{7}{c}{\say{The hail caused bad damage to my garden.} \emph{or} \say{The hail damaged my garden badly.}} \\
\bottomrule
\end{tabular}

\caption{Transliterated Persian with a gloss and translation from \citet{karimi2011separability}, annotated in a Persian-specific schema. The light verb construction \form{\emph{latme zadan}} (\say{to damage}) has been spread across the sentence. Multiword constructions like this are a challenge for word-level tagging schemata.}
\label{fig:light_verb_construction}
\end{figure*}

UniMorph's development is more centralized and pipelined.%
\footnote{This centralization explains why UniMorph tables exist for only 49 languages, or 50 when counting the Norwegian Nynorsk and Bokm\aa{}l writing forms separately.}
Inflectional paradigms are scraped from Wiktionary, annotators map positions in the scraped data to MSDs, and the mapping is automatically applied to all of the scraped paradigms. Because annotators handle languages they are familiar with (or related ones), realization of the schema is also done on a language-by-language basis. Further, the scraping process does not capture lexical aspects that are not inflected, like noun gender in many languages. The schema permits inclusion of these details; their absence is an artifact of the data collection process. Finally, UniMorph records only exist for nouns, verbs, and adjectives, though the schema is broader than these categories.

For these reasons, we treat the corpora as imperfect realizations of the schemata. Moreover, we contend that ambiguity in the schemata leave the door open to allow for such imperfections. With no strict guidance, it's natural that annotators would take different paths. Nevertheless, modulo annotator disagreement, we assume that within a particular corpus, one word form will always be consistently annotated.

Three categories of annotation difficulty are missing values, language-specific attributes, and multiword expressions.

\paragraph{Missing values} In both schemata, irrelevant attributes are omitted for words to which they do not pertain. For instance, an English verb is not labeled \tag{\textbf{Gender}=NULL}; the \tag{\textbf{Gender}} attribute is simply excluded from the annotation, making the human-readable representations compact. Unfortunately, in both resources, even relevant attributes are intentionally omitted. A verb's positiveness, activeness, or finiteness can be taken as implicit, and it will be omitted arbitrarily on a language-by-language basis. For instance, in our example in \autoref{tab:annotations} only UD tags Spanish finite verbs: \tag{VerbForm=Fin}. Not only UniMorph makes such elisions: we note that \emph{neither} resource marks verb forms as active---an action entirely permitted by the schemata.
 This is one source of discrepancy, both between the projects and across languages within a project, but it is straightforward to harmonize. 

\paragraph{Language-specific attributes}
\phantomsection{} \label{sec:lgspec}
UD records a set of features that are kept language-specific, including \tag{\textbf{Position}} in Romanian, \tag{\textbf{Dialect}} in Russian, and \tag{\textbf{NumValue}} in Czech and Arabic.\footnote{The complete list is at \url{http://universaldependencies.org/v2/features.html\#inventory-of-features-that-will-stay-language-specific}} UniMorph has (potentially infinite) language-specific features \tag{LgSpec1}, \tag{LgSpec2}, \ldots, which are sparsely used but opaque when encountered. For instance, \tag{LgSpec1} in Spanish distinguishes between the two (semantically identical) forms of the imperfect subjunctive: the \form{-se} and \form{-ra} forms (e.g.\ \say{\emph{estuviese}} and \say{\emph{estuviera}} from \say{\emph{estar}} \say{to be}). UD does not annotate the forms differently. If a language has multiple language-specific attributes, their order is not prescribed by the UniMorph schema, and separate notes that explain the use of such tags must accompany datasets.

\paragraph{Multiword expressions} A final imperfection is how to represent multiword constructions. Both UD and UniMorph are word-level annotations, espousing what has alternately been called the \term{lexical integrity principle}~\citep{chomsky1970remarks, bresnan1995lexical} or \term{word-based morphology}~\citep{aronoff1976word, aronoff2007beginning, spencer1991morphological}. Unfortunately, not all morphological manifestations occur at the level of individual words. The Farsi (Persian) \term{light verb construction} illustrates the deficiency \citep[see][]{karimi2011separability}. Farsi expresses many actions by pairing a light verb (one with little meaning) with a noun that gives a concrete meaning. The example in \autoref{fig:light_verb_construction} uses the light verb construction \say{\emph{latme zadan}} (\say{to damage}). The parts of the verb construction are separated in the sentence, seeming to require a morphosyntactic parse. When attempting to annotate these constructs, neither schema provides guidance. In languages where these occur, language-specific decisions are made. It should be noted that multiword expressions are a general challenge to natural language processing, not specifically morphology \citep{sag2002multiword}.

\section{A Deterministic Conversion} \label{sec:conversion}

In our work, the goal is not simply to translate one schema into the other, but to translate one \emph{resource} (the imperfect manifestation of the schema) to match the other. The differences between the schemata and discrepancies in annotation mean that the transformation of annotations from one schema to the other is not straightforward.

Two naive options for the conversion are a lookup table of MSDs and a lookup table of the individual attribute-value pairs which comprise the MSDs. The former is untenable: the table of all UD feature combinations (including null features, excluding language-specific attributes) would have \SI{2.445e17} entries. Of course, most combinations won't exist, but this gives a sense of the table's scale. Also, it doesn't leverage the factorial nature of the annotations: constructing the table would require a massive duplication of effort. On the other hand, attribute-value lookup lacks the flexibility to show how a pair of values interacts. Neither approach would handle language- and annotator-specific tendencies in the corpora.

Our approach to converting UD MSDs to UniMorph MSDs begins with the attribute-value lookup, then amends it on a language-specific basis. Alterations informed by the MSD and the word form, like insertion, substitution, and deletion, increase the number of agreeing annotations. They are critical for work that examines the MSD monolithically instead of feature-by-feature \citep[e.g.][]{belinkov2017neural, cotterell2017cross}: Without exact matches, converting the individual tags becomes hollow.

Beginning our process, we relied on documentation of the two schemata to create our initial, language-agnostic mapping of individual values. This mapping has \(140\) pairs in it. Because the mapping was derived purely from the schemata, it is a useful approximation of how well the schemata match up. We note, however, that the mapping does not handle idiosyncrasies like the many uses of \say{dative} or features which are represented in UniMorph by argument templates: possession and ergative--absolutive argument marking. The initial step of our conversion is using this mapping to populate a proposed UniMorph MSD.

As shown in \autoref{sec:results}, the initial proposal is often frustratingly deficient. Thus we introduce the post-edits. To concoct these, we looked into UniMorph corpora for these languages, compared these to the conversion outputs, and then sought to bring the conversion outputs closer to the annotations in the actual UniMorph corpora. When a form and its lemma existed in both corpora, we could directly inspect how the annotations differed. Our process of iteratively refining the conversion implies a table which exactly maps any combination of UD MSD and its related values (lemma, form, etc.) to a UniMorph MSD, though we do not store the table explicitly.

Some conversion rules we've created must be applied before or after others. These sequential dependencies provide conciseness. Our post-editing procedure operates on the initial MSD hypothesis as follows:

\begin{enumerate}
    \item First, we collect all arguments relating to a possessor or an ergative--absolutive language's argument agreement, because UniMorph represents both categories as a single templatic value.
    \item We discard any values that UniMorph doesn't annotate for a particular part of speech, like gender and number in French verb participles, or German noun genders. 
    \item We make MSD additions when they are unambiguously implied by the resources, like \tag{PFV} to accompany \tag{PST} in Spanish \say{pasado simple}, but \tag{PST} to accompany \tag{IPFV} in Spanish \say{pasado continuo}.
    \item We also incorporate fixes using information outside of the MSD like the \tag{LgSpec1} tag for Spanish's \form{-ra} forms, as described in \autoref{sec:lgspec}, and other language-specific corrections, like mapping the various dative cases to the cross-lingually comparable case annotations used in UniMorph.
\end{enumerate}

\paragraph{What we left out} We did, however, reject certain changes that would increase harmony between the resources. Usually, this decision was made when the UniMorph syntax or tagset was not obeyed, such as in the case of made-up tags for Basque arguments (instead of the template mentioned above) or the use of idiopathic colons (:) instead of semicolons (;) as separators in Farsi. Other instances were linguistically motivated. UD acknowledges Italian imperatives, but UniMorph does not have any in its table. We could largely alter these to have subjunctive labels, but to ill effect. A third reason to be conservative in our rules was cases of under-specification: If a participle is not marked as past or present in UD, but both exist in UniMorph, we could unilaterally assign all to the majority category and increase recall. This would pollute the data with fallacious features, so we leave these cases under-specified. In other words, we do not add new values that cannot be unequivocally inferred from the existing data.

\paragraph{Output} The Universal Dependencies data are presented in the CoNLL-U format.\footnote{\url{http://universaldependencies.org/format.html}} Each sentence is represented in tabular form to organize annotations like lemmas, parts of speech, and dependencies of each word token. The MSDs are held in a column called \texttt{FEATS}. Our MSD conversion tool produces a CoNLL-U file whose \texttt{FEATS} column now contains a UniMorph-style MSD. For more straightforward interface with UniMorph, the feature bundle includes the part of speech tag. As the \texttt{POS} column of the CONLL-U file is preserved, this can easily be stripped from the \texttt{FEATS} column, depending on use case.

\paragraph{Why not a learned mapping?} One can imagine learning the UniMorph MSD corresponding to a UD dataset's MSD by a set-to-set translation model like IBM Model~1 \citep{brown1993mathematics}. Unfortunately, statistical (and especially neural) machine translation generalizes in unreliable ways. Our goal is a straightforward, easily manipulable and extensible conversion that prioritizes correctness over coverage.

\section{Experiments}

We evaluate our tool on two tasks:
\begin{description}
\item[Intrinsic assessment:] Once we convert UD MSDs to UniMorph MSDs, how many of the converted ones are attested in UniMorph's paradigm tables.
\item[Extrinsic assessment:] Whether performance on a downstream task is comparable when using pre- and post-conversion MSDs.
\end{description}
To be clear, our scope is limited to the schema conversion. Future work will explore NLP tasks that exploit both the created token-level UniMorph data and the existing type-level UniMorph data.

\paragraph{Data} We draw our input data from the UD v2.1 treebanks \citep{nivre2017universal}. When multiple treebanks exist for a language, we select the one with a basic name, e.g.\ \say{Spanish} instead of \say{Spanish-AnCora}. We leave the construction of additional converters to future work, and we invite the community to participate in designing the mappings for all UD treebanks. UniMorph modifies its language packs individually instead of offering versioned releases. Our UniMorph lookup tables are the latest versions at the time of writing.\footnote{As of 19 June 2018, the latest modification to a UniMorph language resource was to Finnish on 3 August 2017.} There are 31 languages which possess both a UD and a UniMorph corpus.

\subsection{Intrinsic evaluation} We transform all UD data to the UniMorph. We compare the simple lookup-based transformation to the one with linguistically informed post-edits on all languages with both UD and UniMorph data. We then evaluate the recall of MSDs without partial credit. 

\paragraph{Calculating recall} Because the UniMorph tables only possess annotations for verbs, nouns, adjectives, or some combination, we can only examine performance for these parts of speech. We consider two words to be a match if their form and lemma are present in both resources. Syncretism allows a single surface form to realize multiple MSDs (Spanish \form{\emph{mandaba}} can be first- or third-person), so we define success as the computed MSD matching \emph{any} of the word's UniMorph MSDs. This gives rise to an equation for recall: of the word--lemma pairs found in both resources, how many of their UniMorph-converted MSDs are present in the UniMorph tables?

\paragraph{Why no held-out test set?} Our problem here is not a learning problem, so the question is ill-posed. There is no \emph{training} set, and the two resources for a given language make up a test set. The quality of our model---the conversion tool---comes from how well we encode prior knowledge about the relationship between the UD and UniMorph corpora.

\subsection{Extrinsic evaluation} If the UniMorph-converted treebanks perform differently on downstream tasks, then they convey different information. This signals a failure of the conversion process. As a downstream task, we choose morphological tagging, a critical step to leveraging morphological information on new text. 

We evaluate taggers trained on the transformed UD data, choosing eight languages randomly from the intersection of UD and UniMorph resources. We report the macro-averaged F1 score of attribute-value pairs on a held-out test set, with official train/validation/test splits provided in the UD treebanks. As a reference point, we also report tagging accuracy on those languages' untransformed data.

We use the state-of-the-art morphological tagger of \citet{malaviya2018neural}. It is a factored conditional random field with potentials for each attribute, attribute pair, and attribute transition. The potentials are computed by neural networks, predicting the values of each attribute jointly but not monolithically. Inference with the potentials is performed approximately by loopy belief propagation. We use the authors' hyperparameters.

We note a minor implementation detail for the sake of reproducibility. The tagger exploits explicit guidance about the attribute each value pertains to. The UniMorph schema's values are globally unique, but their attributes are not explicit. For example, the UniMorph \tag{Masc} denotes a masculine gender. We amend the code of \citeauthor{malaviya2018neural} to incorporate attribute identifiers for each UniMorph value.

\section{Results} \label{sec:results}

\begin{table}
    \centering
    \begin{tabular}{l S[table-format=5.2] S[table-format=5.2] }
    \toprule
        Language & {CSV} & {Post-editing} \\
    \midrule
        Ar & 0.00 & {-} \\
        Bg & 34.61 & 87.88 \\
        Ca & 23.23 & 99.78 \\
        Cs & 0.48 & 81.71 \\
        Da & 1.55 & 4.70 \\
        De & 17.20 & 60.81 \\
        En & 42.17 & 90.10 \\
        Es & 17.20 & 97.86 \\
        Eu & 0.00 & 0.00 \\
        Fa & 0.00 & {-} \\
        Fi & 59.19 & 92.81 \\
        Fr & 18.61 & 99.20 \\
        Ga & 0.41 & 0.41 \\
        He & 4.08 & 46.61 \\
        Hi & 0.00 & {-} \\
        Hu & 15.46 & 24.94 \\
        It & 22.32 & 94.89 \\
        La & 11.73 & 64.25 \\
        Lt & 0.00 & {-} \\
        Lv & 0.17 & 90.58 \\
        Nb & 2.11 & 38.88 \\
        Nl & 12.12 & 12.12 \\
        Nn & 2.40 & 40.21 \\
        Pl & 7.70 & 88.17 \\
        Pt & 20.11 & 99.34 \\
        Ro & 0.00 & 25.16 \\
        Ru & 0.00 & {-} \\
        Sl & 37.57 & 90.27 \\
        Sv & 13.20 & 83.44 \\
        Tr & 0.00 & 65.14 \\
        Uk & 4.06 & 96.45 \\
        Ur & 0.00 & 55.72 \\

    \bottomrule
    \end{tabular}
    \caption{Token-level recall when converting Universal Dependencies tags to UniMorph tags. CSV refers to the lookup-based system. Post-editing refers to the proposed method.}
    \label{tab:recall}
\end{table}

\begin{table}
    \centering
    \begin{tabular}{l S[table-format=5.2] S[table-format=5.2]}
    \toprule
        Language & {{UD F1}} & {{UniMorph F1}} \\
    \midrule
        Da & 90.58 & 92.59 \\
        Es & 78.31 & 96.44 \\
        Fi & 93.78 & 94.98 \\
        Lv & 84.20 & 86.94 \\
        Pt & 95.57 & 95.77 \\
        Ru & 89.89 & 89.95 \\
        Bg & 95.54 & 95.79 \\
        Sv & 92.39 & 93.83 \\
    \bottomrule
    \end{tabular}
    \caption{Tagging F1 using UD sentences annotated with either original UD MSDs or UniMorph-converted MSDs}
    \label{tab:tagging}
\end{table}

We present the intrinsic task's recall scores in \autoref{tab:recall}. Bear in mind that due to annotation errors in the original corpora (like the \say{\emph{vas}} example from \autoref{sec:resources}), the optimal score is not always \(100\%\). Some shortcomings of recall come from irremediable annotation discrepancies. Largely, we are hamstrung by differences in choice of attributes to annotate. When one resource marks gender and the other marks case, we can't infer the gender of the word purely from its surface form. The resources themselves would need updating to encode the relevant morphosyntactic information. Some languages had a very low number of overlapping forms,\footnote{Fewer than \(250\) overlapping form--lemma pairs. The other languages had overlaps in the thousands.} and no tag matches or near-matches between them: Arabic, Hindi, Lithuanian, Persian, and Russian. A full list of observed, irremediable discrepancies is presented alongside the codebase.

There are three other transformations for which we note no improvement here. Because of the problem in Basque argument encoding in the UniMorph dataset---which only contains verbs---we note no improvement in recall on Basque. Irish also does not improve: UD marks gender on nouns, while UniMorph marks case. Adjectives in UD are also underspecified. The verbs, though, are already correct with the simple mapping. Finally, with Dutch, the UD annotations are impoverished compared to the UniMorph annotations, and missing attributes cannot be inferred without external knowledge.

For the extrinsic task, the performance is reasonably similar whether UniMorph or UD; see \autoref{tab:tagging}. A large fluctuation would suggest that the two annotations encode distinct information. On the contrary, the similarities suggest that the UniMorph-mapped MSDs have similar content. We recognize that in every case, tagging F1 increased---albeit by amounts as small as \(0.16\) points. This is in part due to the information that is lost in the conversion. UniMorph's schema does not indicate the type of pronoun (demonstrative, interrogative, etc.), and when lexical information is not recorded in UniMorph, we delete it from the MSD during transformation. On the other hand, UniMorph's atomic tags have more parts to guess, but they are often related. (E.g.\ \tag{Ipfv} always entails \tag{Pst} in Spanish.) Altogether, these forces seem to have little impact on tagging performance.

\section{Related Work}

The goal of a tagset-to-tagset mapping of morphological annotations is shared by the Interset project \citep{zeman2008reusable}. Interset decodes features in the source corpus to a \emph{tag interlingua}, then encodes that into target corpus features. (The idea of an interlingua is drawn from machine translation, where a prevailing early mindset was to convert to a universal representation, then encode that representation's semantics in the target language. Our approach, by contrast, is a direct flight from the source to the target.) Because UniMorph corpora are noisy, the encoding from the interlingua would have to be rewritten for each target. Further, decoding the UD MSD into the interlingua cannot leverage external information like the lemma and form. 

The creators of HamleDT sought to harmonize dependency annotations among treebanks, similar to our goal of harmonizing across resources \citep{zeman2014hamledt}. The treebanks they sought to harmonize used multiple diverse annotation schemes, which the authors unified under a single scheme.

\citet{petrov2011universal} present mappings into a coarse, \say{universal} part of speech for 22 languages. Working with POS tags rather than morphological tags (which have far more dimensions), their space of options to harmonize is much smaller than ours.

Our extrinsic evaluation is most in line with the paradigm of \citet{wisniewski2017systematic} (and similar work therein), who compare syntactic parser performance on UD treebanks annotated with two styles of dependency representation. Our problem differs, though, in that the dependency representations express different relationships, while our two schemata vastly overlap. As our conversion is lossy, we do not appraise the learnability of representations as they did.

In addition to using the number of extra rules as a proxy for harmony between resources, one could perform cross-lingual projection of morphological tags \citep{drabek2005induction, kirov2017rich}. Our approach succeeds even without parallel corpora.

\section{Conclusion and Future Work}

We created a tool for annotating Universal Dependencies CoNLL-U files with UniMorph annotations. Our tool is ready to use off-the-shelf today, requires no training, and is deterministic. While under-specification necessitates a lossy and imperfect conversion, ours is interpretable. Patterns of mistakes can be identified and ameliorated.

The tool allows a bridge between resources annotated in the Universal Dependencies and Universal Morphology (UniMorph) schemata. As the Universal Dependencies project provides a set of treebanks with token-level annotation, while the UniMorph project releases type-level annotated tables, the newfound compatibility opens up new experiments. A prime example of exploiting token- and type-level data is \citet{tackstrom2013token}. That work presents a part-of-speech (POS) dictionary built from Wiktionary, where the POS tagger is also constrained to options available in their type-level POS dictionary, improving performance. Our transformation means that datasets are prepared for similar experiments with morphological tagging. It would also be reasonable to incorporate this tool as a subroutine to UDPipe \citep{straka2017tokenizing} and Udapi \citep{popel2017udapi}. We leave open the task of converting in the opposite direction, turning UniMorph MSDs into Universal Dependencies MSDs.

Because our conversion rules are interpretable, we identify shortcomings in both resources, using each as validation for the other. We were able to find specific instances of incorrectly applied UniMorph annotation, as well as specific instances of cross-lingual inconsistency in both resources. These findings will harden both resources and better align them with their goal of universal, cross-lingual annotation.

\section*{Acknowledgments}

We thank Hajime Senuma and John Sylak-Glassman for early comments in devising the starting language-independent mapping from Universal Dependencies to UniMorph.

\bibliography{ud_unimorph}
\bibliographystyle{acl_natbib_nourl}

\end{document}